\documentclass{IEEEcsmag}

\usepackage[colorlinks,urlcolor=blue,linkcolor=blue,citecolor=blue]{hyperref}
\usepackage{subfig}
\usepackage{upmath}
\usepackage{tcolorbox}
\usepackage{todonotes}

\jvol{XX}
\jnum{XX}
\paper{8}
\jmonth{May/June}
\jname{Multimedia}
\pubyear{2021}

\begin{document}

\title{Why Accuracy Is Not Enough: The Need for Consistency in Object Detection}

\author{Caleb Tung, Abhinav Goel, Fischer Bordwell, Nick Eliopoulos, Xiao Hu, Yung-Hsiang Lu}
\affil{\{tung3, goel39, fbordwel, neliopou, hu440, yunglu\}@purdue.edu - Purdue University}

\author{George K. Thiruvathukal}
\affil{gkt@cs.luc.edu - Loyola University Chicago}

\begin{abstract}
Object detectors are vital to many modern computer vision applications.
However, even state-of-the-art object detectors are not perfect.
On two images that look similar to human eyes, the same detector can make different predictions because of small image distortions like camera sensor noise and lighting changes. 
This problem is called inconsistency.
Existing accuracy metrics do not properly account for inconsistency, and similar work in this area only targets improvements on artificial image distortions.
Therefore, we propose a method to use non-artificial video frames to measure object detection consistency over time, across frames.
Using this method, we show that the consistency of modern object detectors ranges from 83.2\% to 97.1\% on different video datasets from the Multiple Object Tracking Challenge.
We conclude by showing that applying image distortion corrections like .WEBP Image Compression and Unsharp Masking can improve consistency by as much as 5.1\%, with no loss in accuracy.
\end{abstract}

\maketitle

\chapterinitial{Much of modern computer vision relies on object detectors.} An object detector is a Deep Neural Network (DNN) that takes an image as the input and then identifies the locations and types of objects found in that image. Across scientific disciplines, object detectors are increasingly ubiquitous. From electronic package sorting in e-commerce to collision detection in traffic monitoring, from remote sensing in low-orbit satellites to automated MRI screening in the fight against cancer: object detectors are driving an entire frontier of technology.
With so many critical applications, object detectors need to be consistently accurate. Modern object detectors use different architectures (e.g., single-shot, R-CNN, etc.) and training methods (e.g., multitask loss, neural architecture search, etc.) to achieve state-of-the-art accuracy on popular image datasets like Microsoft COCO (Common Objects in Context) \cite{lin_microsoft_2014}.


\section{The Consistency Problem}
Even though object detectors are carefully tested for accuracy, this article observes that \textit{consistency} is also a valuable metric that receives less attention in literature. As we discuss later, common image datasets make it challenging to test for consistency.
\begin{figure}[h]
    \centering
    \subfloat[]{
        \includegraphics[width=0.46\linewidth]{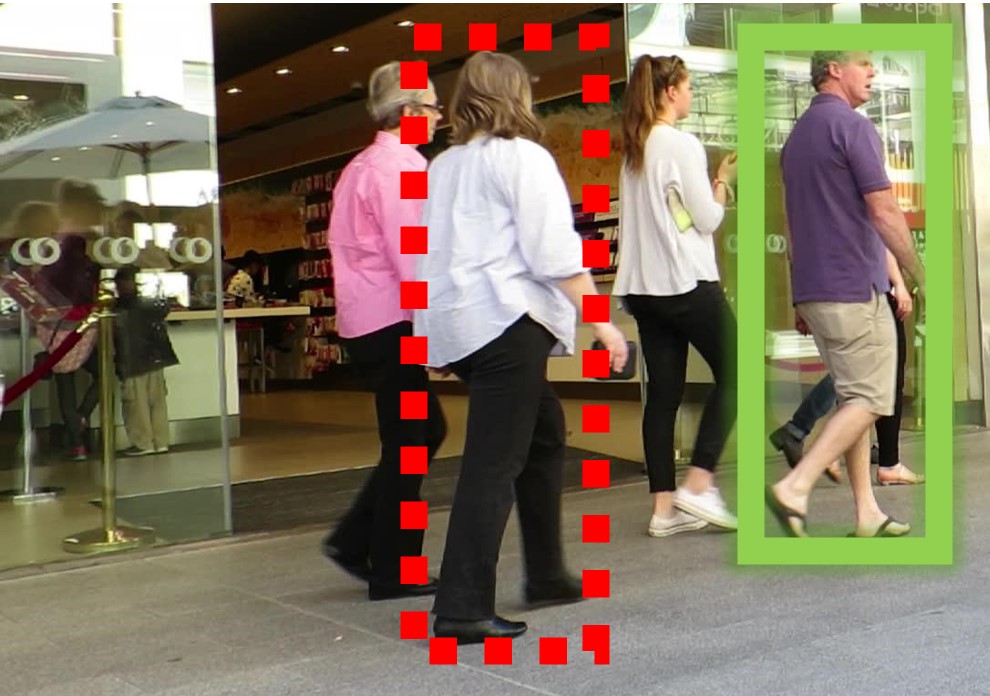}
    }
    \subfloat[]{
        \includegraphics[width=0.46\linewidth]{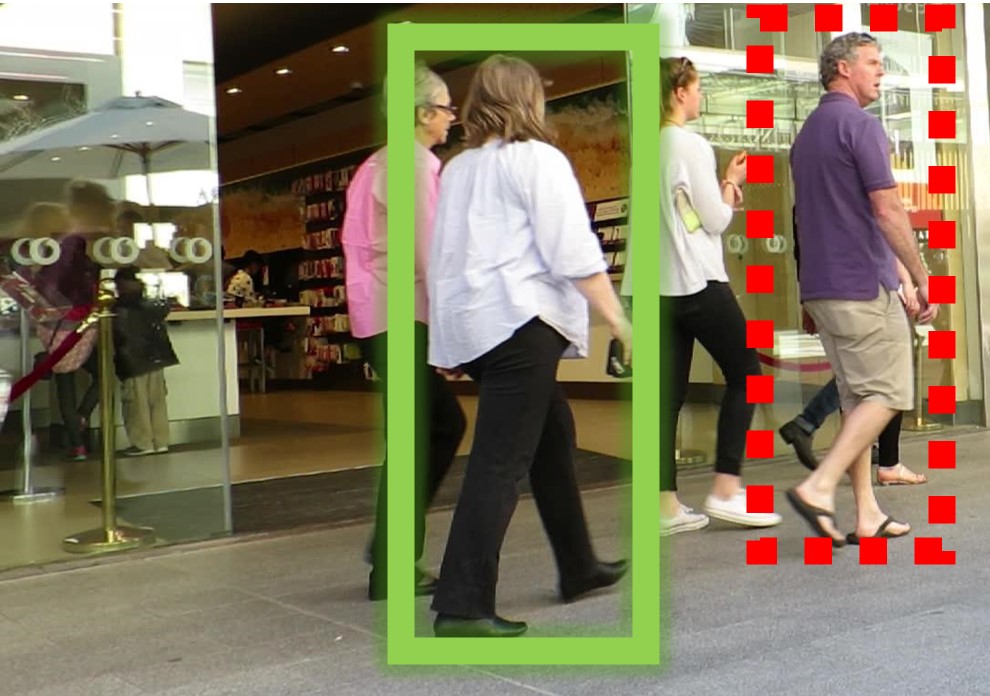}
    }
    \caption{\small State-of-the-art object detector Mask-RCNN is inconsistent on two images taken 0.03s apart, even though both images look alike. In (a), the woman is missed (red dashed-line box) while the man is detected (green solid box). In (b), the reverse is true. (All other people are detected correctly in both images, giving an average 3/4 accuracy in both images.) \normalsize}
    \label{fig:maskrcnn-inconsistent}
\end{figure}

Accuracy typically measures how often an object detector is correct \textit{on average}. This information is partially deficient because it does not capture the variation in an object detector's performance when input images are similar.
Since accuracy is reported as an average, there could be multiple ways an object detector achieves a given accuracy, some of which may be less desirable. For example, in Fig. \ref{fig:maskrcnn-inconsistent}, a state-of-the-art object detector (Mask-RCNN \cite{he_mask_2018}) detects three out of four people per image. The accuracy is the same \textit{on average} (3/4), yet the detector behaves inconsistently: it misses a different person in each image.


Inconsistent behavior becomes cause for concern in vision applications that demand strict performance guarantees. For example, a collision prevention algorithm with 95\% accuracy that misses the same 5\% of objects allows one to investigate the cause of the 5\% error more easily, because the defective behavior is \textit{consistent}. However, an algorithm that behaves inconsistently, missing different objects across each image in the test, is much harder to troubleshoot.

\begin{figure}[h]
    \centering
    \subfloat[]{
        \includegraphics[width=0.46\linewidth]{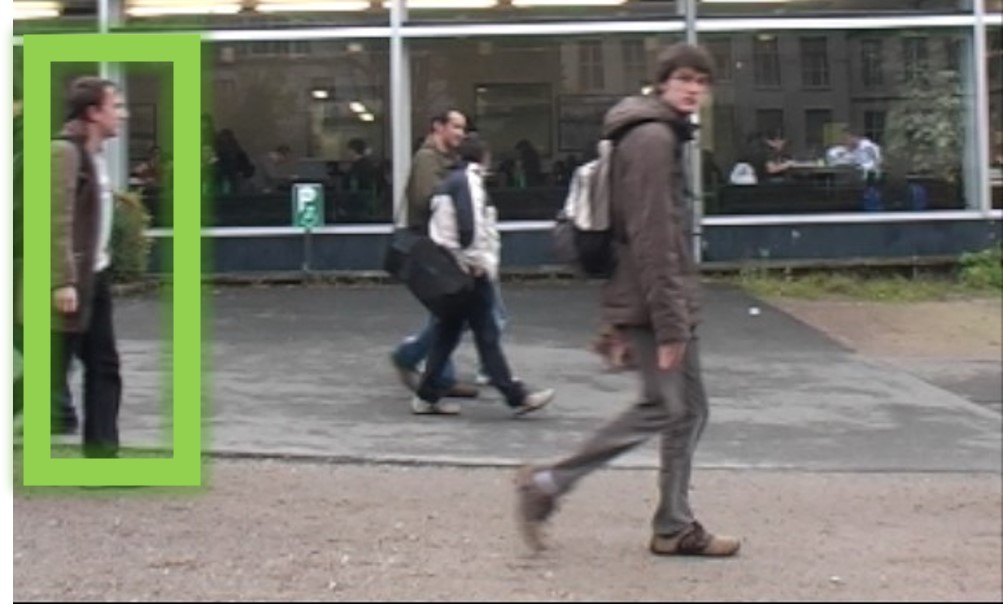}
    }
    \subfloat[]{
        \includegraphics[width=0.46\linewidth]{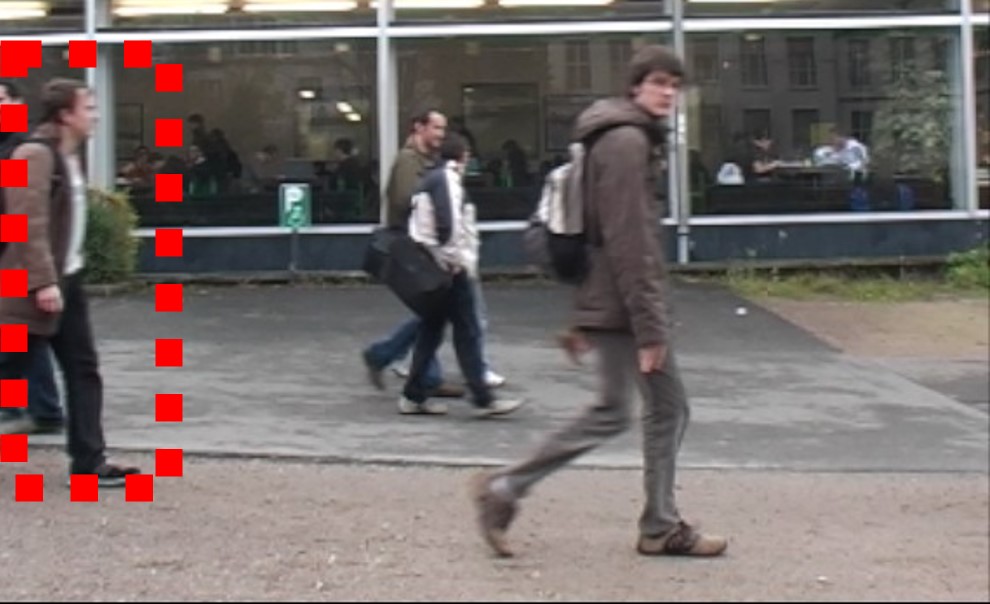}
    }
    
    \subfloat[]{
        \includegraphics[width=0.46\linewidth]{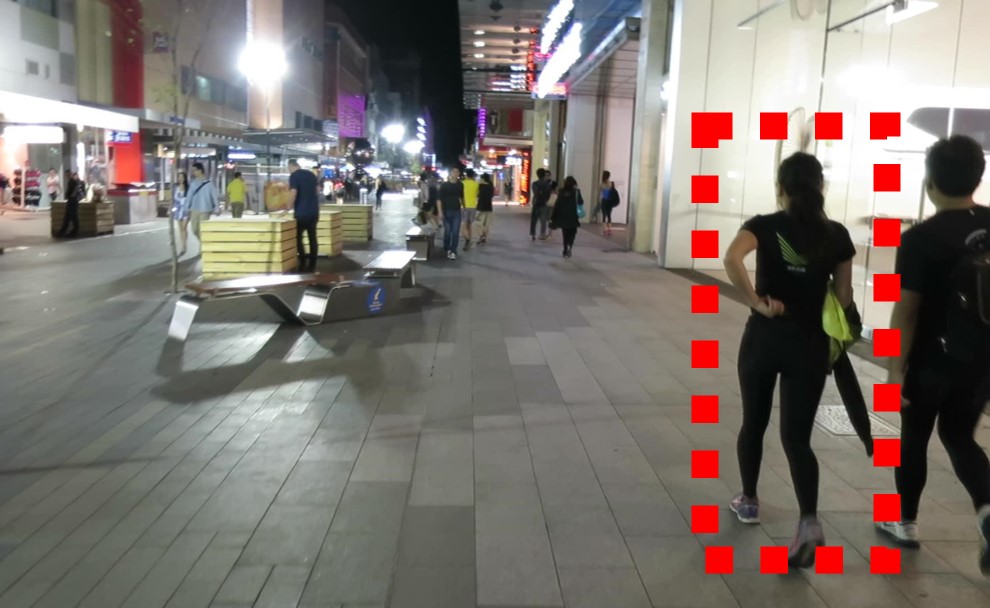}
    }
    \subfloat[]{
        \includegraphics[width=0.46\linewidth]{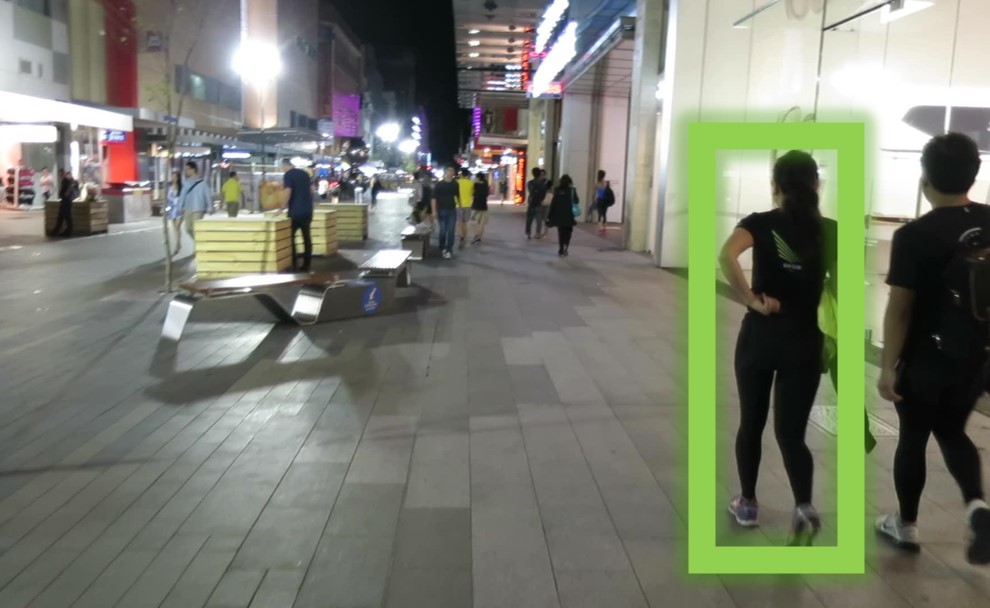}
    }
    
    \subfloat[]{
        \includegraphics[width=0.46\linewidth]{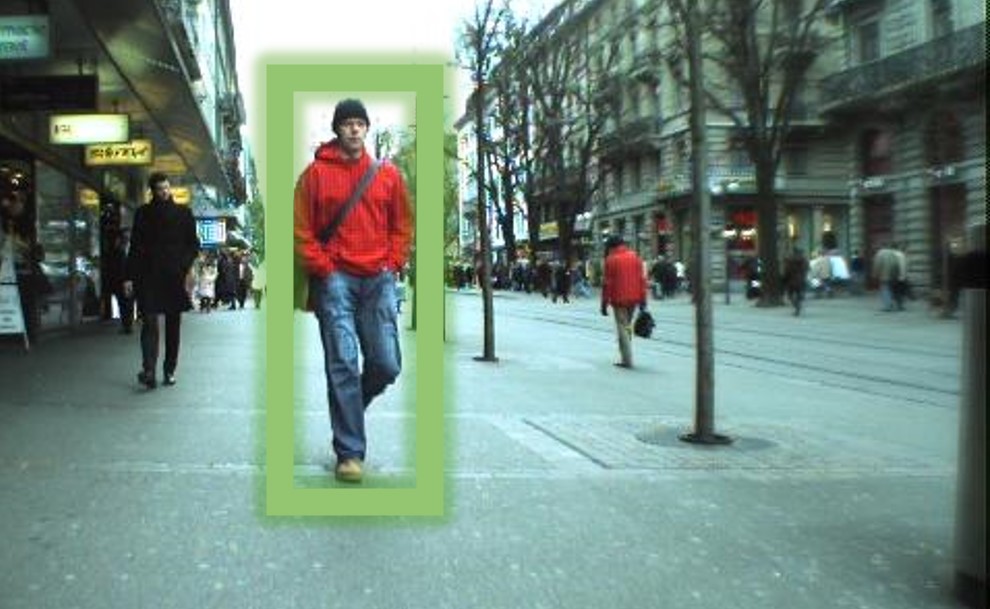}
    }
    \subfloat[]{
        \includegraphics[width=0.46\linewidth]{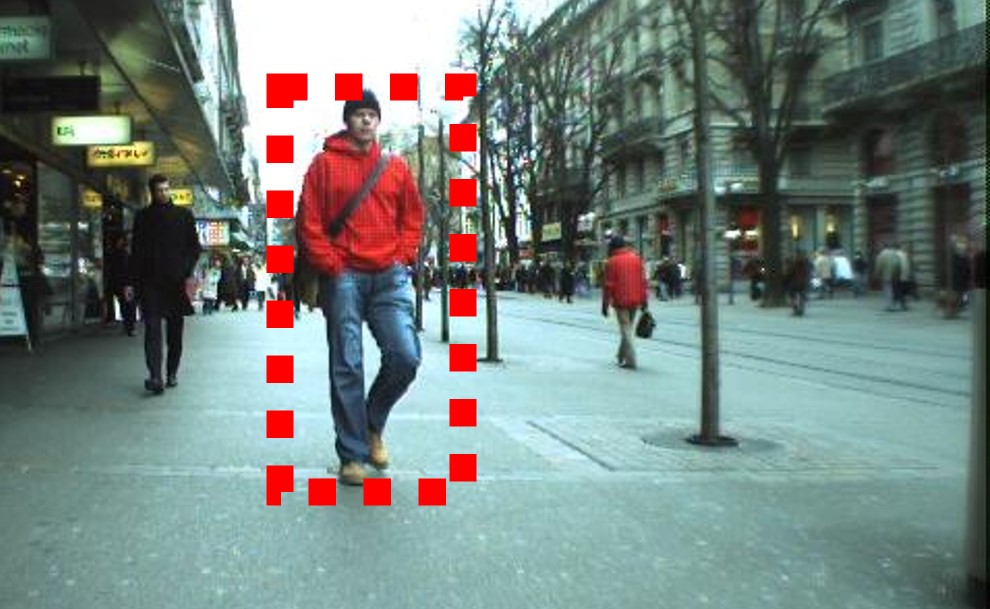}
    }
    \caption{\small Different object detectors behave inconsistently on images that look visually similar. Left images are taken 0.03s before the right. Red, dashed-line boxes are missed detections, green solid boxes are correct detections. (a)-(b) Faster-RCNN. (c)-(d) RetinaNet. (e)-(f) SSD. \normalsize}
    \label{fig:object-detectors-inconsistent}
\end{figure}

This article investigates object detector \textit{consistency} as a method to augment existing accuracy metrics. Consistency measures the difference in predictions from an object detector across \textit{similar} images. We explore different methods to quantify consistency, ultimately choosing a metric that tracks a detector’s behavior on time-series images from the MOT Challenge \cite{leal-taixe_motchallenge_2015}---a dataset originally intended to benchmark object tracking. As shown in Fig. \ref{fig:object-detectors-inconsistent}, we find that state-of-the-art detectors (Mask-RCNN, Faster-RCNN \cite{ren_faster_2015}, RetinaNet \cite{lin_focal_2017}, SSD \cite{liu_ssd_2016}) exhibit inconsistent behavior.  
In our experiments, we observe up to an average of 17\% inconsistent detections.
We evaluate methods to improve consistency and present a selection of methods (lossy image compression, gamma boosting, etc.) that successfully raise consistency by up to 5\%.

\section{What accuracy measures}
Accuracy is determined by comparing the predictions of an object detector against the ground truth in a dataset of images. Standard accuracy metrics like mean Average Precision (mAP) usually consider two factors simultaneously: (1) how much of the ground truth the detector successfully predicts, and (2) how many of the detector’s predictions were incorrect. To get perfect accuracy, each of the predictions an object detector makes must be correct: it cannot blanket the image with guesses.

The mAP accuracy metric is defined in terms of \textit{precision} and \textit{recall} using true/false positives and negatives (see Tab. \ref{tab:positive-negative}). Precision and recall are defined in Eqn. \ref{eqn:precision} and \ref{eqn:recall}, respectively. mAP is calculated for a given image dataset by integrating the area under the precision-versus-recall curve with respect to recall (Eqn. \ref{eqn:accuracy}).

\begin{table}[h]
\caption{Terminology used for calculating object detection accuracy.}
\begin{tabular}{|r|l|l|}
\hline
\textbf{}                                                                & \textbf{Object detected}                                      & \textbf{Object not detected}                                  \\ \hline
\textbf{Object exists}                                                   & \begin{tabular}[c]{@{}l@{}}True Positive\\ (TP)\end{tabular}  & \begin{tabular}[c]{@{}l@{}}False Negative\\ (FN)\end{tabular} \\ \hline
\textbf{\begin{tabular}[c]{@{}r@{}}Object does\\ not exist\end{tabular}} & \begin{tabular}[c]{@{}l@{}}False Positive\\ (FP)\end{tabular} & \begin{tabular}[c]{@{}l@{}}True Negative\\ (TN)\end{tabular}  \\ \hline
\end{tabular}
\label{tab:positive-negative}
\end{table}

\small
\begin{equation}
    \texttt{precision } p = \frac{TP}{TP+FP}
    \label{eqn:precision}
\end{equation}

\begin{equation}
    \texttt{recall } r = \frac{TP}{TP+FN}
    \label{eqn:recall}
\end{equation}

\begin{equation}
    \texttt{accuracy } = \int_{0}^{1} p(r) dr
    \label{eqn:accuracy}
\end{equation}
\normalsize

\section{Accuracy does not measure consistency}
\begin{figure}
    \centering
    \subfloat[]{
        \includegraphics[width=0.46\linewidth]{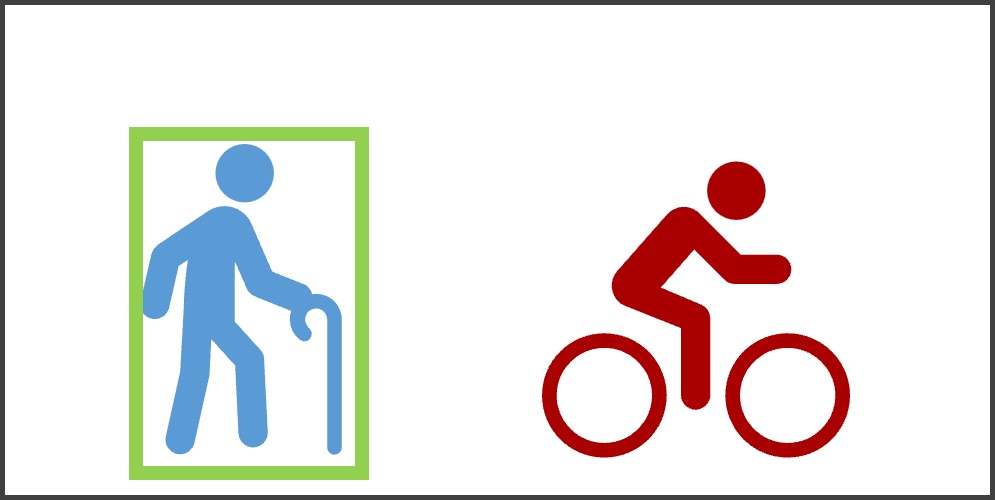}
    }
    \subfloat[]{
        \includegraphics[width=0.46\linewidth]{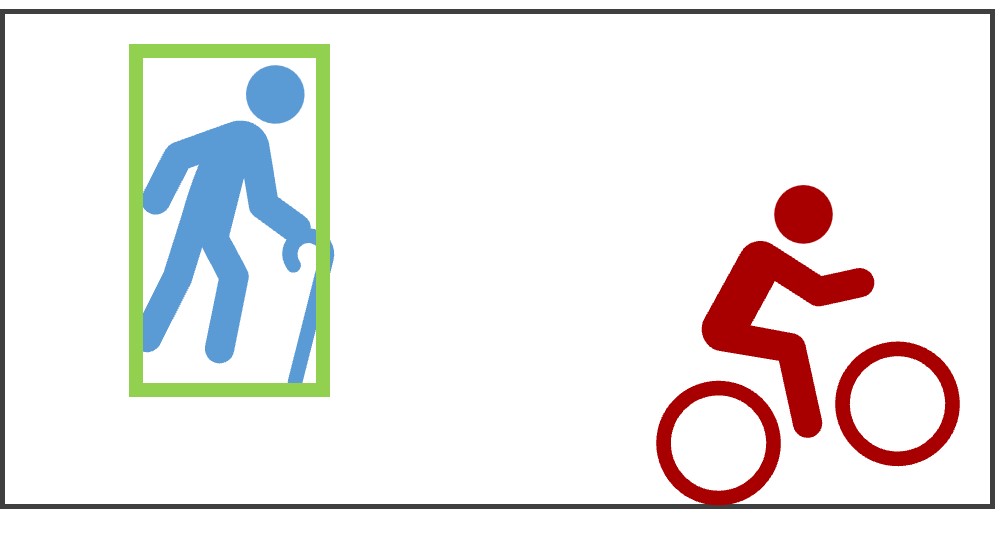}
    }
    
    \subfloat[]{
        \includegraphics[width=0.46\linewidth]{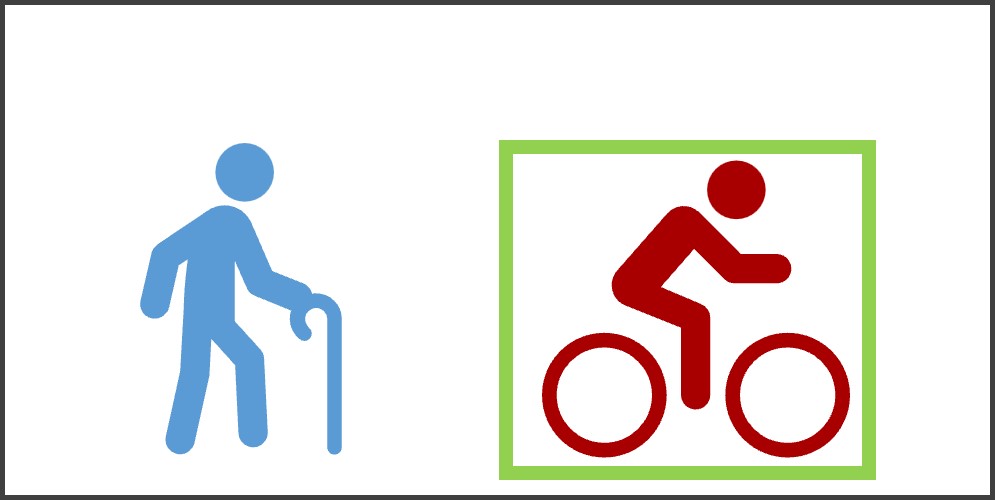}
    }
    \subfloat[]{
        \includegraphics[width=0.46\linewidth]{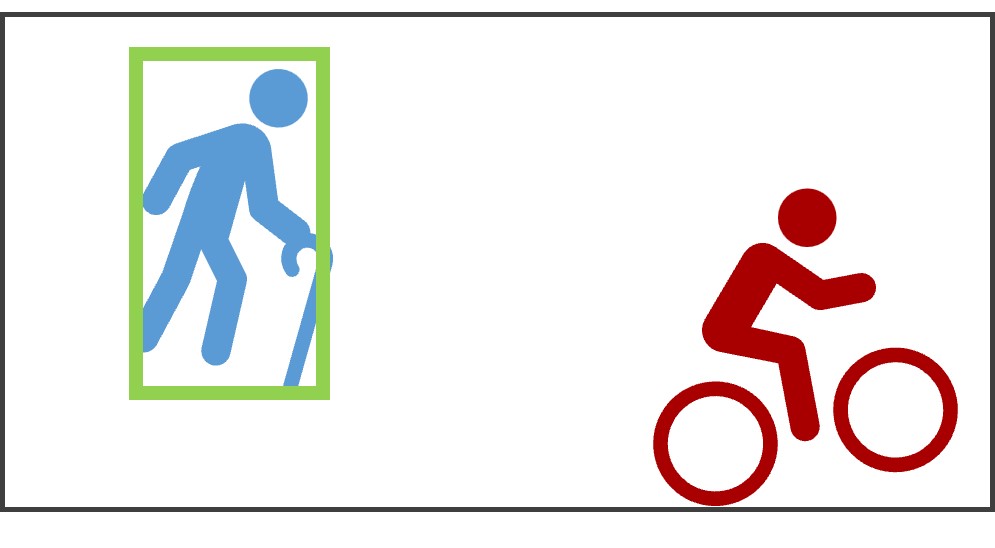}
    }
    
    \caption{\small Accuracy does not capture consistency. For object detections shown in similar-looking images, (a)-(b) has the same average accuracy (i.e. 1/2 correct) as (c)-(d). (a)-(b) has consistent detections (same object detected in both images), but (c)-(d) is inconsistent (different objects detected in both images).\normalsize}
    \label{fig:consistent-vs-inconsistent}
\end{figure}

When accuracy is not 100\%, there may be more to the story (for context, the state-of-the-art Mask-RCNN reaches 63\% mAP on the COCO dataset). Accuracy does not describe whether  the detector is consistent. 
Consider an object detector that achieves 50\% accuracy on a set of images, where each image is slightly different but still contains the same objects. One might hope that the detector would predict consistently as shown in Fig. \ref{fig:consistent-vs-inconsistent}a-b. However, it is still possible for other detector predictions (Fig. \ref{fig:consistent-vs-inconsistent}c-d) to achieve the same 50\% accuracy, albeit inconsistently.

Additionally, simply reporting fine-grained accuracy statistics does not capture consistency. It is true that we can infer more information about the neural network’s consistency by reporting additional statistics like the standard deviation or variance of per-image accuracy across a dataset of similar images. However, reporting the standard deviation still does not reveal the Fig. \ref{fig:consistent-vs-inconsistent}c-d case discussed above, where the detector inconsistently detects and misses \textit{different} objects in each frame even though the \textit{number} of objects remains the same.

As we showed earlier in Fig. \ref{fig:object-detectors-inconsistent}, inconsistent behavior exists even in popular object detectors. Thus, accuracy does not always communicate the full picture of a detector’s performance because it does not explicitly describe consistency. Inconsistency may have severe consequences in applications that involve safety.

\section{Related work}
There is a growing number of studies  to improve computer vision, but they do not  focus on consistency. These efforts can largely be grouped into two categories: (1) adversarial attacks and (2) synthetic image distortions.


\subsection{Adversarial attacks}
\begin{figure}
    \centering
    \subfloat[]{
        \includegraphics[width=0.2\linewidth]{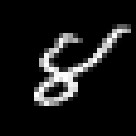}
    }
    \subfloat[]{
        \includegraphics[width=0.2\linewidth]{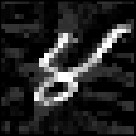}
    }
    \subfloat[]{
        \includegraphics[width=0.2\linewidth]{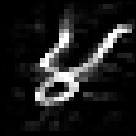}
    }
    \subfloat[]{
        \includegraphics[width=0.2\linewidth]{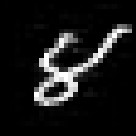}
    }
    
    \caption{\small Similar-looking adversarial samples generated from an image in the MNIST handwriting dataset. (a) Original image. Correctly classified as “8” with 0.9 confidence. (b) Fast-Gradient Sign Method adversarial sample. Wrongly classified as “4” with 0.9 confidence \cite{goodfellow_explaining_2015}. (c) Jacobian Saliency Map Attack adversarial sample. Wrongly classified as “9” with 0.6 confidence \cite{moosavi-dezfooli_universal_2017}. (d) DeepFool adversarial sample. Wrongly classified as “4” with 0.83 confidence \cite{moosavi-dezfooli_deepfool_2016}. 
    \normalsize}
    \label{fig:adversarial-attacks}
\end{figure}

Adversarial attacks present a significant challenge to neural networks. Goodfellow, et al. \cite{goodfellow_explaining_2015} demonstrate that a well-trained image classifier can be tricked into misclassifying an image by slightly perturbing the values of the pixels. In a typical adversarial attack, an algorithm makes minute, calculated adjustments to the pixels of a correctly predicted image until the neural network makes an incorrect prediction. The final image, known as the adversarial sample, appears to  human eyes as very similar to the original image. Several different adversarial samples are shown in Fig. \ref{fig:adversarial-attacks}.

Existing methods to defend against adversarial attacks often involve some combination of (1) including adversarial samples during network training \cite{goodfellow_explaining_2015}, (2) transforming input images into a lower-dimensional space before feeding the neural networks \cite{dorsi_sparse_2020}, and (3) filtering adversarial samples through a custom neural network before they reach the main network \cite{gu_towards_2015}.

\subsection{Synthetic image distortions and data augmentation}
Computer vision models can make incorrect predictions on images from natural sources (i.e., not manipulated for adversarial attacks) \cite{dodge_understanding_2016}. Two images of the same scene can be captured by the same camera less than a second apart, yet the predictions of a neural network on those two visually similar images can differ dramatically. The small differences between the two images are called \textit{image distortions}, and they can be caused by a range of factors, including ambient light level and camera sensor noise \cite{tung_large-scale_2019}.

\begin{figure}
    \centering
    \subfloat[]{
        \includegraphics[width=0.46\linewidth]{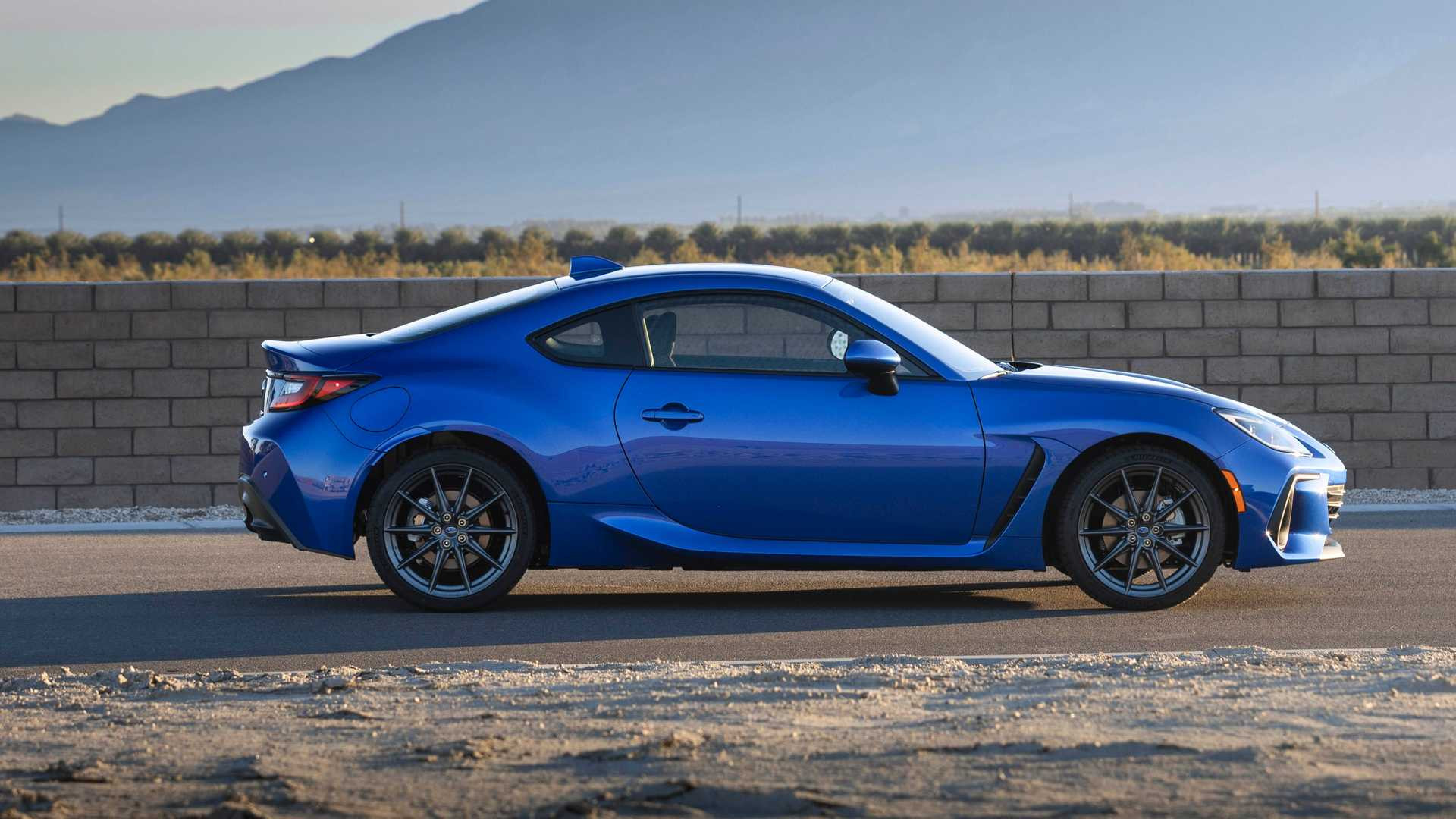}
    }
    \subfloat[]{
        \includegraphics[width=0.46\linewidth]{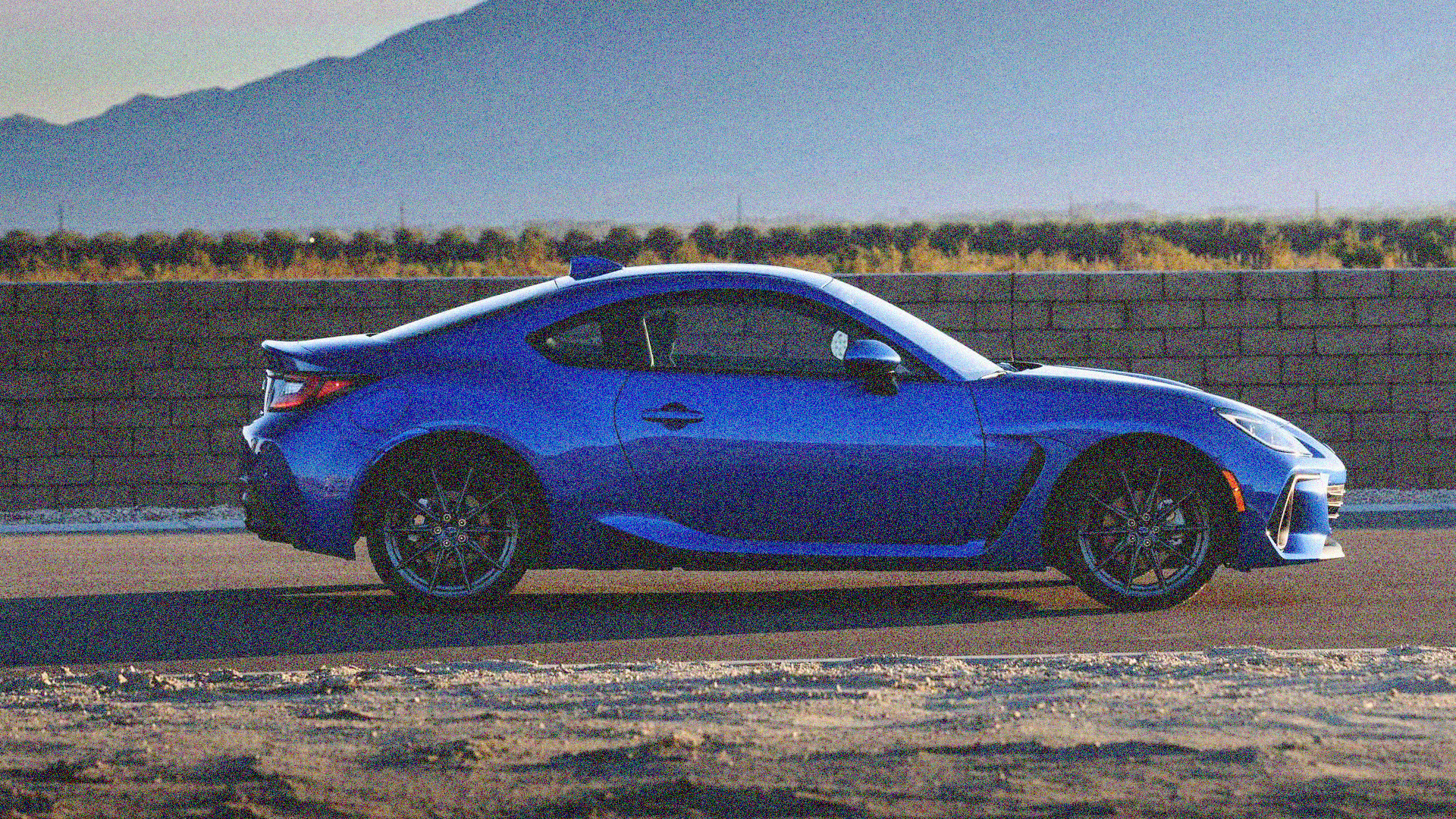}
    }
    
    \subfloat[]{
        \includegraphics[width=0.46\linewidth]{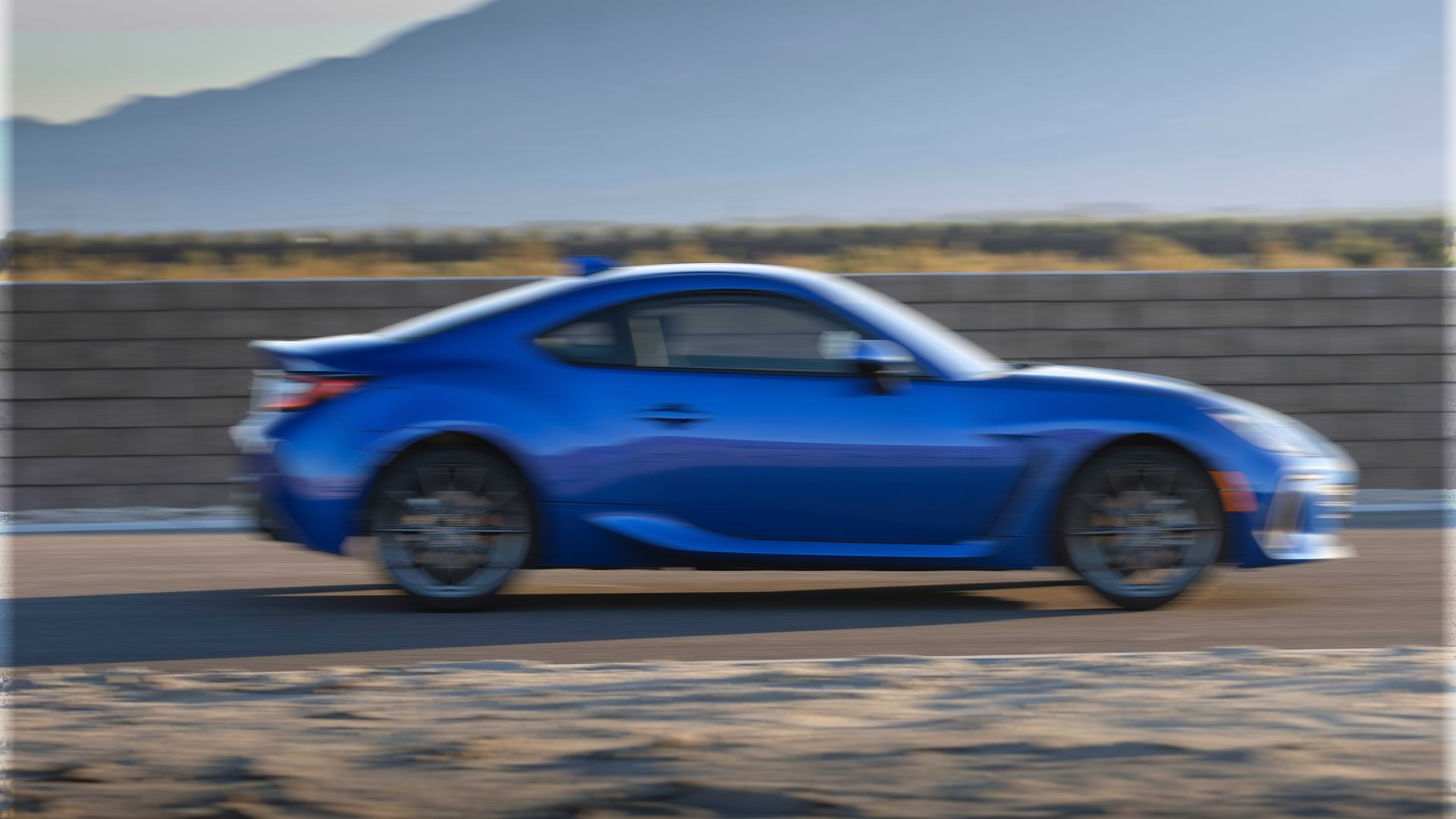}
    }
    \subfloat[]{
        \includegraphics[width=0.46\linewidth]{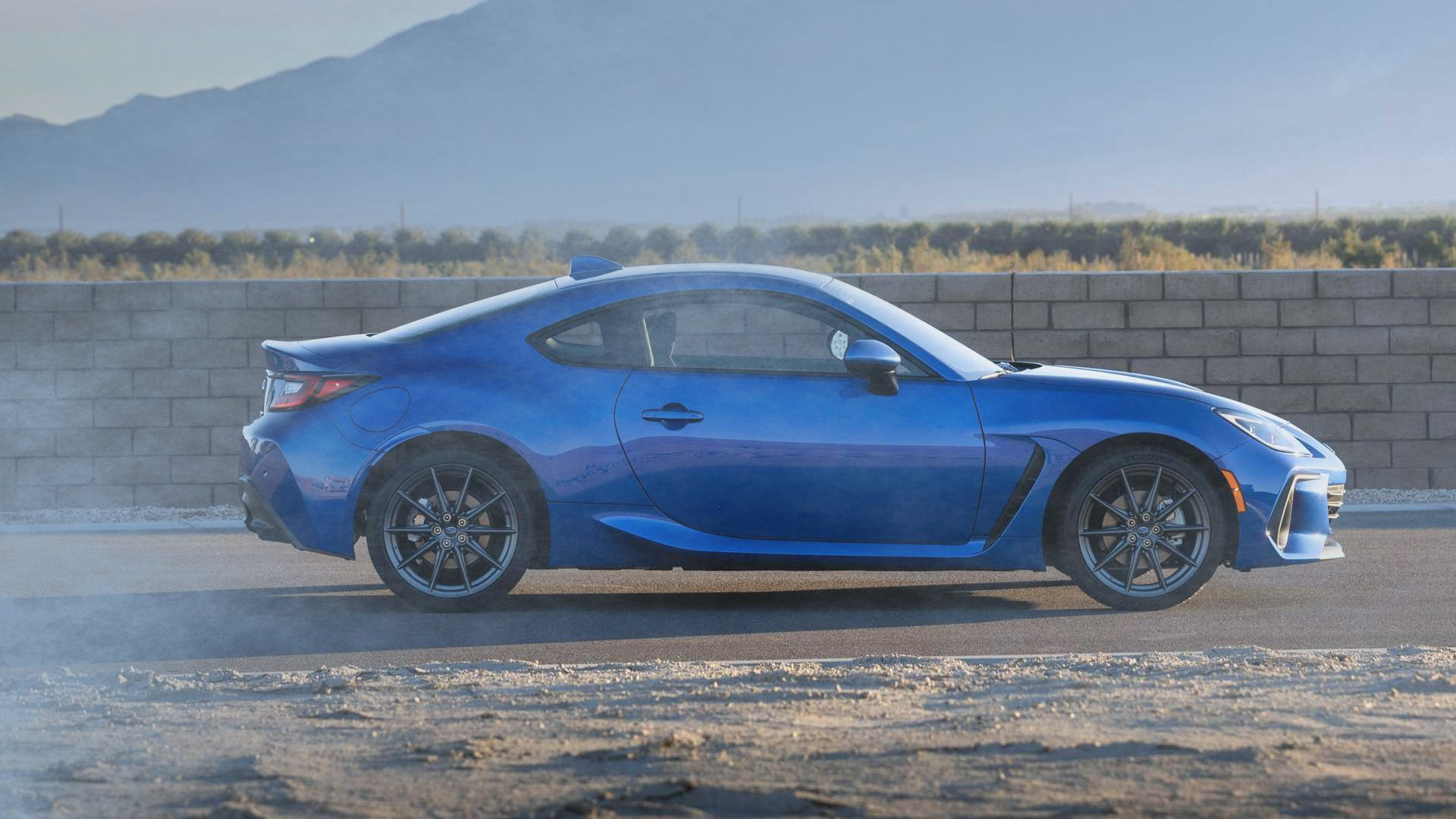}
    }
    
    \caption{\small Examples of synthetic image distortions. (a) Original image. (b) Increased brightness. (c) Motion blur. (d) Artificial fog.\normalsize}
    \label{fig:synthetic-image-distortions}
\end{figure}

To make neural networks more robust against image distortions, data augmentation is widely used. The networks are trained on datasets that are 
modified, or synthetically distorted, to emulate the natural distortions. 
Common synthetic image distortions (shown in Fig. \ref{fig:synthetic-image-distortions}) include perturbing the pixels with Gaussian noise, adding artificial motion blur, adjusting color saturation and brightness, and even adding computer-generated fog, snow, and rain. \cite{dodge_understanding_2016}

The discussed previous work uses
only accuracy as the metrics,
without explicitly measuring consistency. Although some publications \cite{zhang_making_2019, gu_using_2019} explore ways to measure the robustness of a neural network beyond accuracy, they neither concretely define consistency nor propose solutions. This article aims to do both.
\section{Considerations when measuring consistency}
In this article, we define consistency as a metric of how differently an object detector behaves on similar images:

\begin{center}
\textit{If images appear similar to the human eye, an object detector should consistently detect the same objects.}
\end{center}

Since accuracy does not always quantify consistency, we now discuss the considerations needed to properly capture consistency.

\subsection{Use Video/Time-Series Data: consistency measurements require similar test images}
\begin{figure}
    \centering
    \subfloat[]{
        \includegraphics[width=0.3\linewidth]{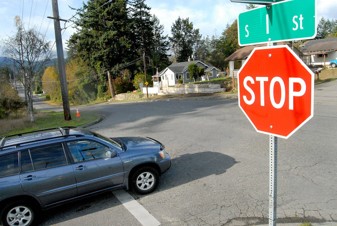}
    }
    \subfloat[]{
        \includegraphics[width=0.3\linewidth]{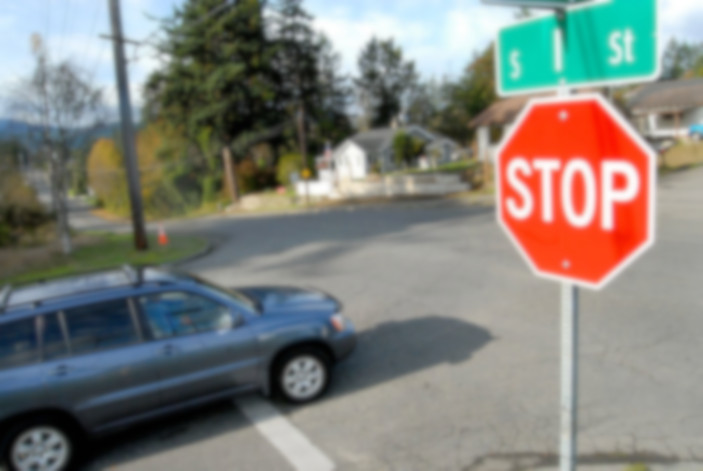}
    }
    \subfloat[]{
        \includegraphics[width=0.3\linewidth]{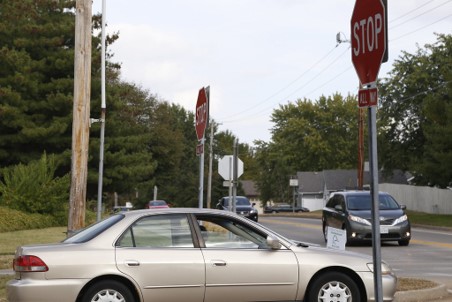}
    }
    
    \caption{\small Consistency is only meaningful when measured across images that look similar (a), (b) (where (b) is taken a second later and camera autofocus is slightly blurred). Consistency is meaningless on images that look different (a), (c). Popular datasets  use images that look different, making them appropriate
    for measuring accuracy, but
    less suited for measuring consistency. \normalsize}
    \label{fig:meaningful-images}
\end{figure}

A meaningful metric for consistency should use input images that are already consistent. Consider Fig. \ref{fig:meaningful-images}a, b. Those pictures appear similar to the human eye, so we would expect consistent performance from an object detector. However, if the images are significantly different (Fig. \ref{fig:meaningful-images}a, c), it is meaningless to use them to make claims about consistency.

As Tung, et al. \cite{tung_large-scale_2019} observe, popular image datasets (e.g., ImageNet, Microsoft COCO) are filled with visually dissimilar images. Thus, those datasets are largely unsuitable for consistency testing. Instead, those authors recommend consecutive frames from videos as a better source of visually similar images.
Although adversarial attacks and artificial image transformations can also be used to generate consistency test data from such popular datasets, Gu, et al. \cite{gu_using_2019} report that such techniques do not well represent image distortions that would occur naturally. Instead, this paper uses consecutive frames from video to test consistency - this way, any inconsistencies would be caused by natural image distortions. 

\begin{figure}
    \centering
    \begin{tcolorbox}[width=\linewidth]
        \section{Quantifying image similarity}
        This article emphasizes the importance of using similar-looking images to test for consistency. We use consecutive frames from videos in the MOT Challenge because such frames are taken up to 30 times per second, ensuring that adjacent frames look similar.
        
        
        Beyond the scope of this article, other popular image similarity measurement methods attempt to relate two images beyond raw pixel value differences. The Structural Similarity Index (SSIM) \cite{wang_image_2004} identifies structural details about the images and then compares the details. This means that SSIM identifies a noisy version of an image as similar to the original, even though raw pixel values are very different. Beyond SSIM, techniques such as the Feature Similarity Index (FSIM) \cite{zhang_fsim_2011} and deep-learning driven comparison extract low-level features to better approximate the way humans compare images.

    \end{tcolorbox}
\end{figure}

\subsection{Use MOT Ground Truth: Consistency measurements need additional labels}
When benchmarking an object detector, bounding box and class ground truth labels are sufficient to report accuracy, but they cannot reveal all inconsistencies. In particular, that ground truth cannot show whether the same objects were detected between two similar images (the above Fig. \ref{fig:consistent-vs-inconsistent}c-d problem): the ground truth is identifier-agnostic. Thus, we need a way to check whether objects from two images are the same, so that consistency can be  measured.

We choose to use per-image object identifier (Object ID) ground truth labels to  keep track of objects during measurement. Each unique object is assigned the same Object ID across the dataset. 



\section{Proposed method of measuring consistency}



Based on our prior discussion of accuracy vs. consistency, we present a method that specifically tracks whether an object detector detects the same objects, given visually similar, time-series images.
Our source of visually similar images is the Multi-Object Tracking (MOT) Challenge \cite{leal-taixe_motchallenge_2015}, consisting of high-quality videos from various datasets.





\begin{figure}[h]
    \centering
    \subfloat[]{
        \includegraphics[width=0.46\linewidth]{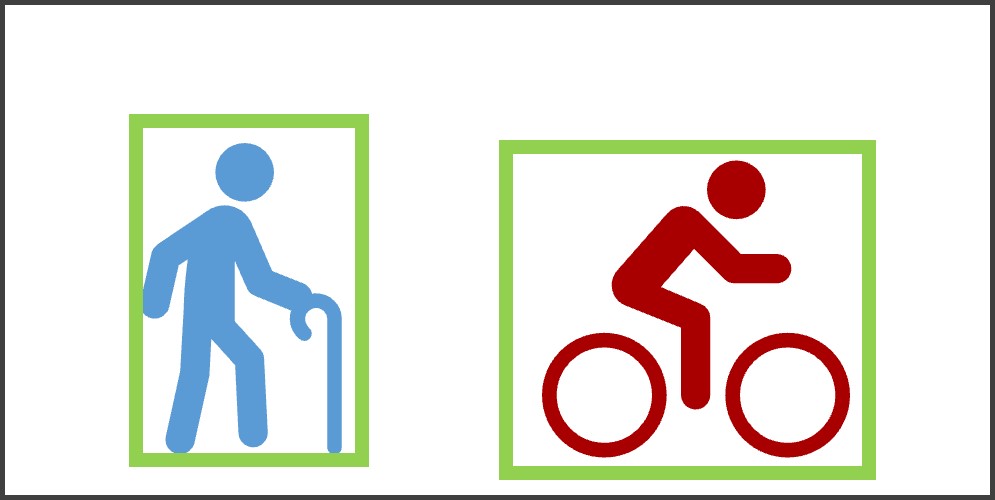}
    }
    \subfloat[]{
        \includegraphics[width=0.46\linewidth]{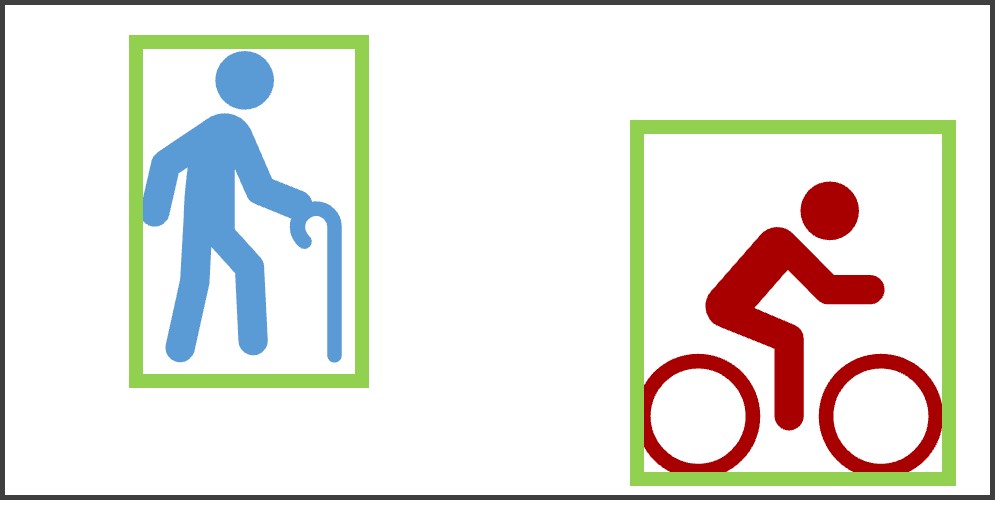}
    }
    
    \subfloat[]{
        \includegraphics[width=0.46\linewidth]{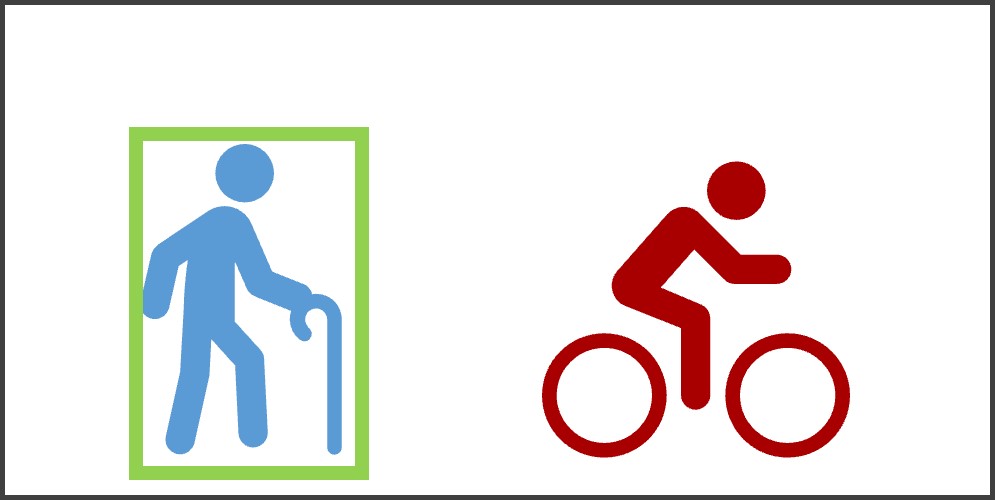}
    }
    \subfloat[]{
        \includegraphics[width=0.46\linewidth]{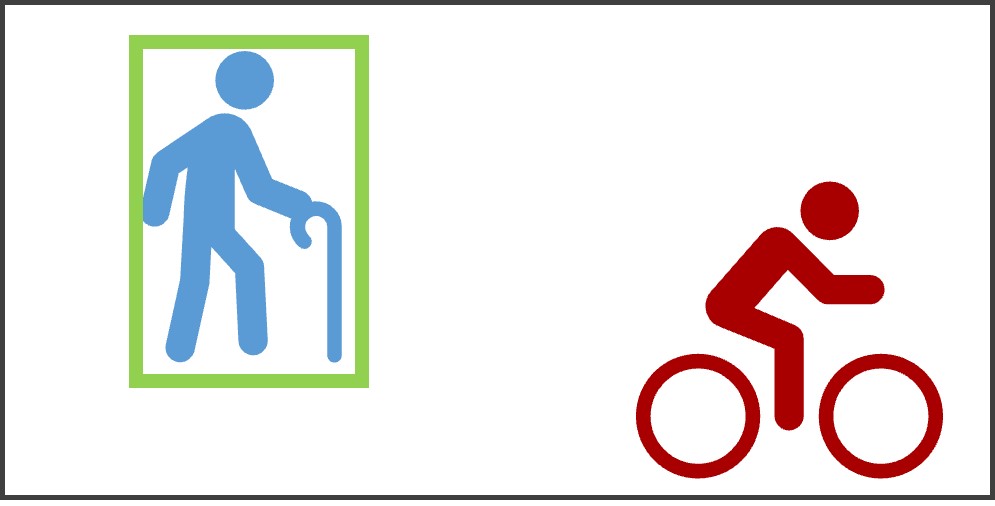}
    }
    
    \subfloat[]{
        \includegraphics[width=0.46\linewidth]{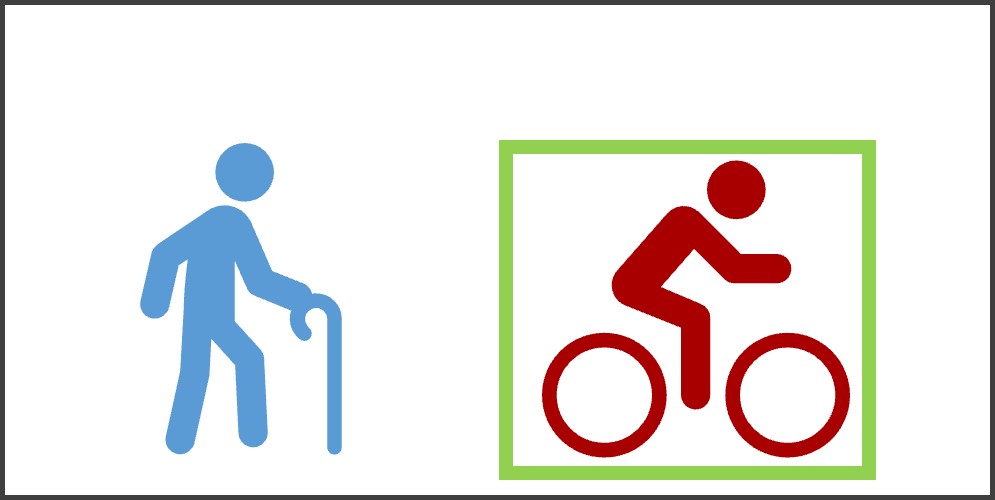}
    }
    \subfloat[]{
        \includegraphics[width=0.46\linewidth]{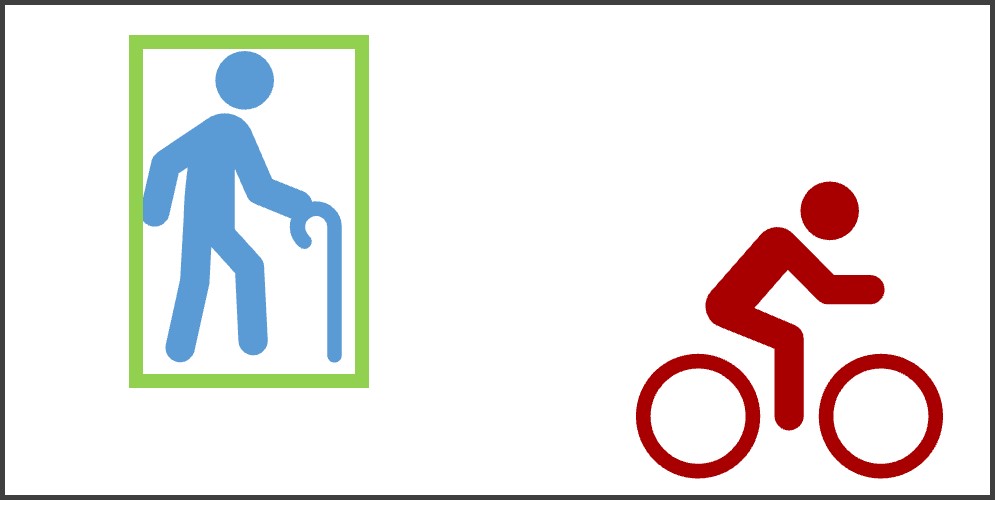}
    }
    
    \subfloat[]{
        \includegraphics[width=0.46\linewidth]{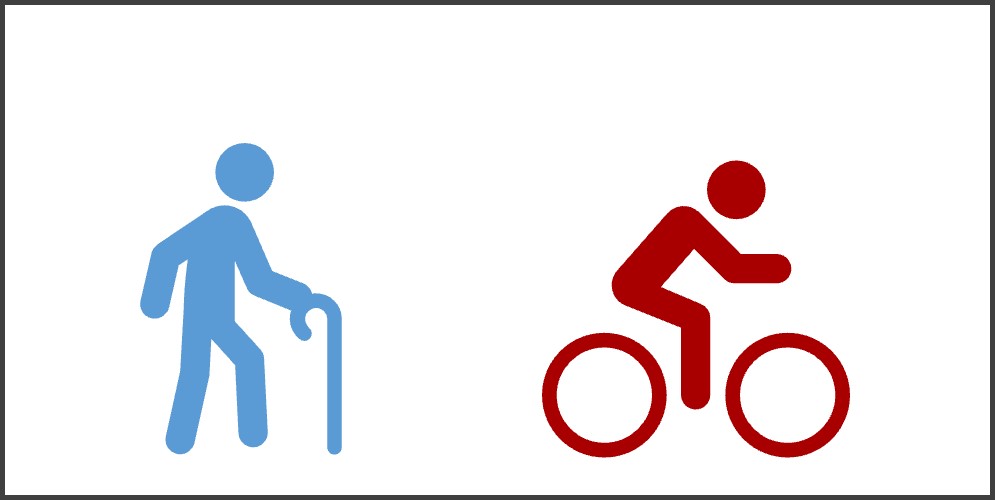}
    }
    \subfloat[]{
        \includegraphics[width=0.46\linewidth]{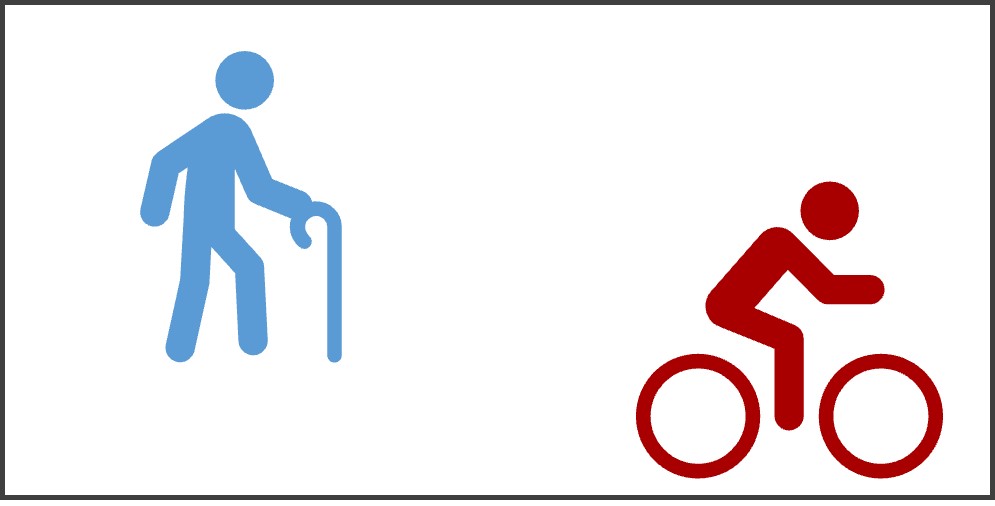}
    }
    \caption{\small Consistency tracks whether an object is detected in one image and missed in another similar-looking image. This complements accuracy measurements. (a)-(b) 100\% accurate (all objects correctly detected in both images), 100\% consistent (nothing was missed in one image and detected in another). (c)-(d) 50\% accurate (only one of two objects detected in each image), 100\% consistent (nothing was missed in one image and detected in another). (e)-(f) 50\% accurate (only one of two objects detected in each image), 0\% consistent (any objects detected in one image was missed in the other). (g)-(h) 0\% accurate (nothing detected), 100\% consistent (nothing  was missed in one image and detected in another). Improving accuracy does not necessarily imply improving consistency and vice versa. \normalsize}
    \label{fig:consistency-works-with-accuracy}
\end{figure}

The \textit{pairwise consistency} $C_{i,j}$ refers to the object detector's consistency on a pair of images $I_i$ and $I_j$. It is calculated as shown in Eqn. \ref{eqn:pairwise-consistency}. If the detector is perfectly consistent, $C_{i,j}$ is 1. If it is entirely inconsistent, then $C_{i,j}$ is 0. As shown in Fig. \ref{fig:consistency-works-with-accuracy}, consistency looks to capture whether objects were detected in one image and missed in another.

\begin{equation}
    C_{i,j}=\frac{|G_i \cap G_j |-|M_{i,j}|-|M_{j,i}|}{|G_i \cap G_j|}
    \label{eqn:pairwise-consistency}
\end{equation}


\begin{figure}[h]
    \centering
    \subfloat[]{
        \includegraphics[width=0.46\linewidth]{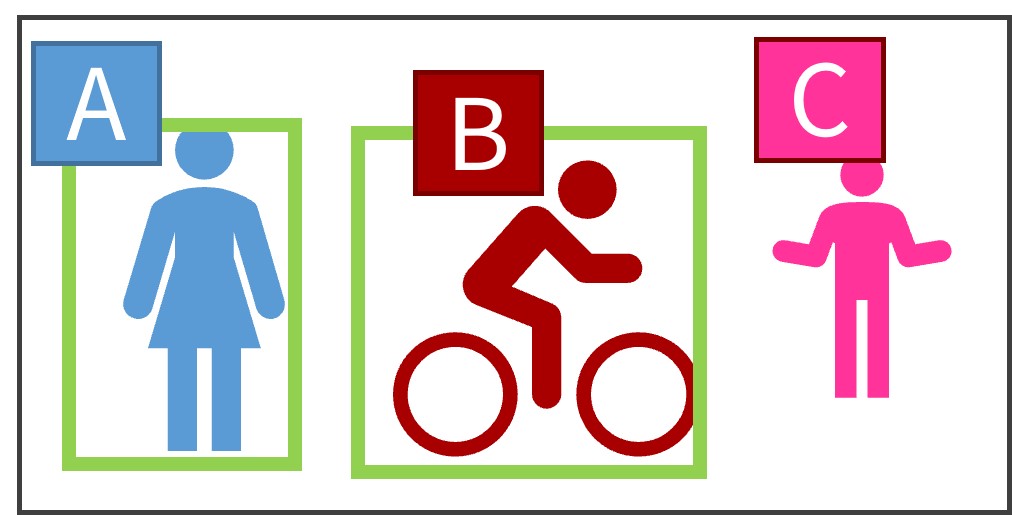}
    }
    \subfloat[]{
        \includegraphics[width=0.46\linewidth]{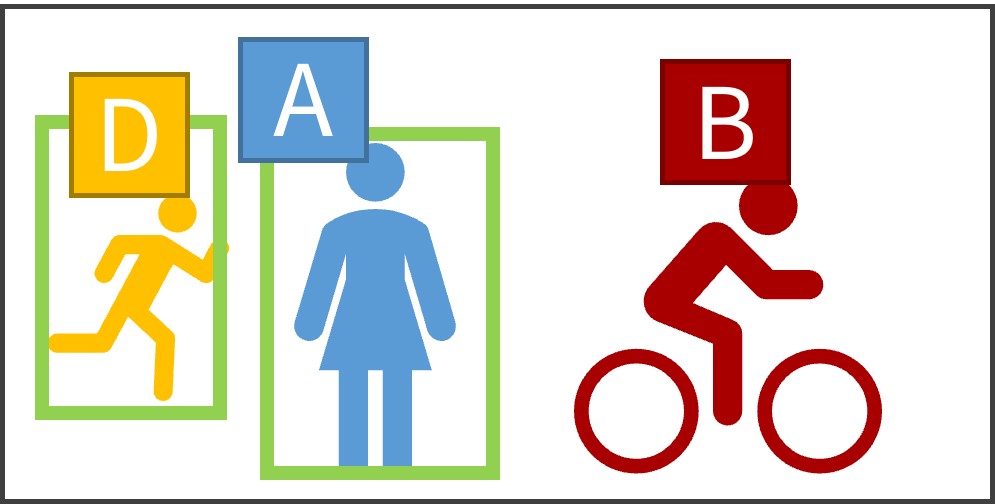}
    }

    \caption{\small Visualization of Eq. \ref{eqn:pairwise-consistency}, where (a) is $I_i$, (b) is $I_j$ and the green boxes indicate object detections. $G_i \cap G_j$ contains objects A and B since they appear in both $I_i, I_j$. Objects C and D are not included in calculations because they do not appear in both images. Because object B is detected in $I_i$ but missed in $I_j$, so $M_{i,j}$ contains object B. No shared boxes are detected in $I_j$ and missed in $I_i$, so $M_{j,i}=\emptyset$. Thus, consistency $C_{i,j}=(2-1-0)/2=0.5$.  \normalsize}
    \label{fig:consistency-explained}
\end{figure}

Eqn. \ref{eqn:pairwise-consistency} is explained using the example in Fig. \ref{fig:consistency-explained}. Fig. \ref{fig:consistency-explained}a is image $I_i$, and Fig. \ref{fig:consistency-explained}b is image $I_j$. $G_i$ is the set of $I_i$'s ground truth Object IDs \{A, B, C\}, and $G_j$ is the set of $I_j$'s ground truth Object IDs \{D, A, B\}. $M_{i,j}, M_{j,i}$ captures the objects that were inconsistently detected as follows: $M_{i,j}$ is the set of ground truth Object IDs that satisfy the following conditions: (1) the ground truth box is present in both images $I_i, I_j$ (i.e. in $G_i \cap G_j$), (2) the object detector detected the object in frame $I_i$, and (3) the object detector missed the object in frame $I_j$. So $M_{i,j}$ is object B, while $M_{j,i}$ is empty.

If an object is present in both images, yet is detected in only one of the images, than consistency should decrease. Taken together, $M_{i,j}, M_{j,i}$ captures consistency decreases in $I_i,I_j$. This also implies that if both $|M_{i,j} |,|M_{j,i} |$=0, the detector can be said to be consistent.



Bounding boxes predicted by the object detector are eligible for consideration in $M_{i,j},M_{j,i}$ calculations only after being filtered through non-maximum suppression (we use the common IoU threshold = 0.5) and a confidence threshold of 0.7 (see sidebar: ``IoU and non-maximum suppression'').

\begin{figure}
    \centering
    \begin{tcolorbox}[width=\linewidth]
        \section{IoU and non-maximum suppression}
        \textit{IoU (Intersection-over-Union)} is a common measurement in object detection, used to determine if two bounding boxes overlap sufficiently to be counted as the same object. It is calculated by dividing the area of two bounding boxes’ overlap by the area of the union of the two boxes. If IoU = 1, then the boxes perfectly overlap. If IoU = 0, the boxes have no overlap. In literature, an IoU threshold of 0.5 is typically used \cite{tung_large-scale_2019} to decide if two bounding boxes sufficiently overlap.
        
        \textit{Non-maximum suppression} is a common application of IoU to filter an object detector’s predictions so that only the “best” ones remain. It finds all overlapping predicted bounding boxes (determined by the IoU threshold) and then filters out the ones that have the same object class and lower confidence scores.
    \end{tcolorbox}
\end{figure}


We measure consistency across a given video $V$ with $N$ frames by measuring pairwise consistency between each pair of adjacent frames in the video, and then averaging the results across all $N-1$ pairs. This is expressed in Eqn. \ref{eqn:overall-consistency}.

\begin{equation}
    C_V=\frac{1}{N-1} \sum_{i=1}^{N-1}C_{i,i+1}
    \label{eqn:overall-consistency}
\end{equation}


As shown earlier in Fig. \ref{fig:consistency-works-with-accuracy}, consistency and accuracy can work together to supply more information about object detection performance than either could on its own. Improving accuracy does not necessarily imply improved consistency and vice versa. Colloquially, one might say that consistency indicates how similarly an object detector behaves on two similar images, while accuracy indicates whether that behavior is desirable (detects everything consistently) or undesirable (misses everything consistently).

Finally, note that we choose to compare bounding box object IDs instead of output feature maps to calculate consistency. This is because comparing feature maps requires arbitrary similarity metrics - selecting an appropriate metric is itself an open problem. Additionally, bounding boxes are already used for accuracy measurements; re-using boxes for consistency will make it more convenient for researchers to take consistency measurements.

\section{Consistency of modern object detectors}
We present the consistency of several state-of-the-art object detectors, showing that they exhibit inconsistent behavior. We demonstrate using highly accurate two-stage object detectors (Faster-RCNN and Mask-RCNN) as well as the faster, but less accurate, single-shot detectors (RetinaNet and SSD). We use Facebook’s official, pretrained models from their \texttt{torchvision} Python package. Measurements are taken on the videos found in the MOT Challenge.


\begin{figure}
    \centering
    \subfloat[]{
        \includegraphics[width=0.96\linewidth]{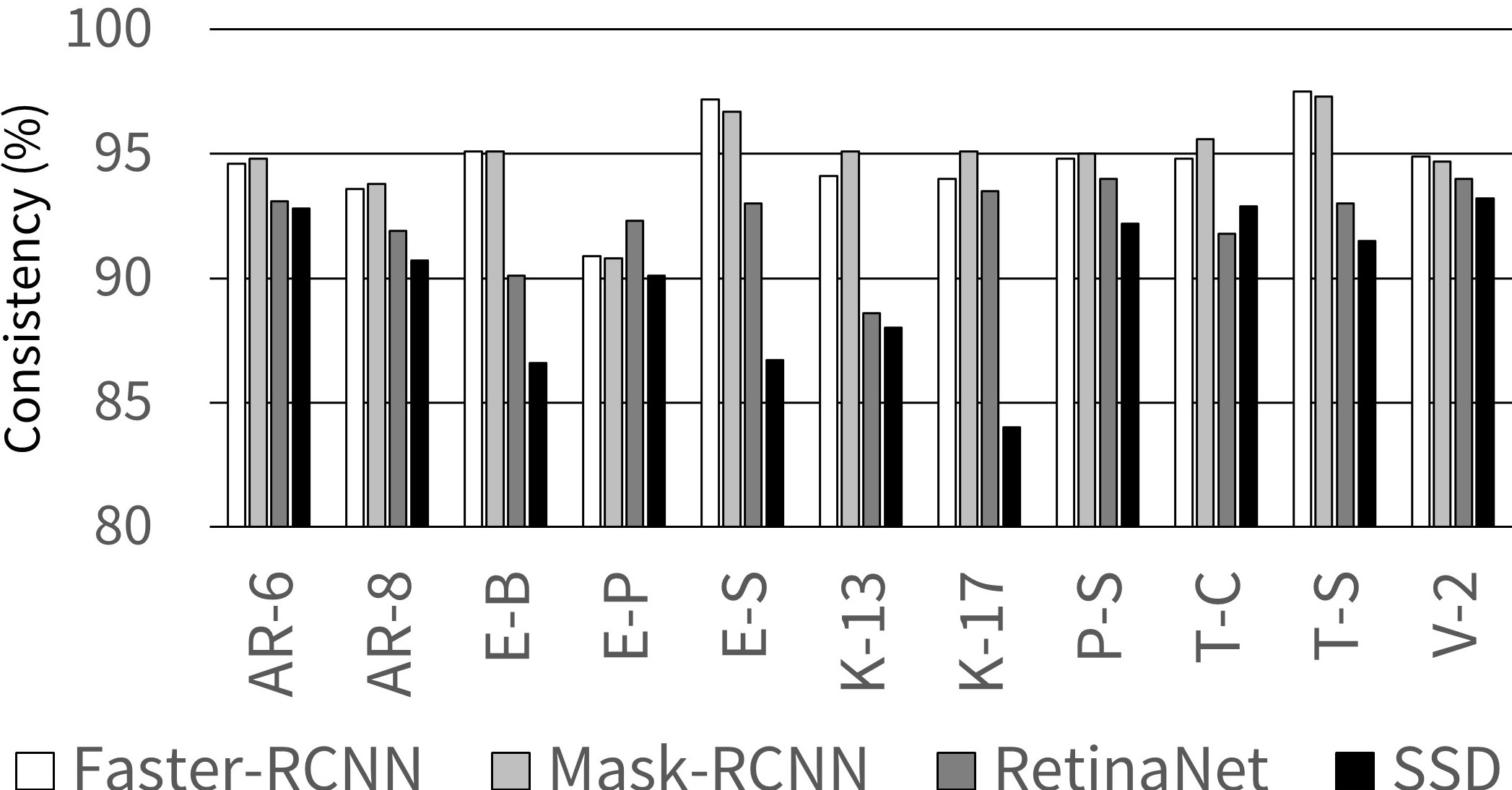}
    }
    
    \subfloat[]{
        \includegraphics[width=0.96\linewidth]{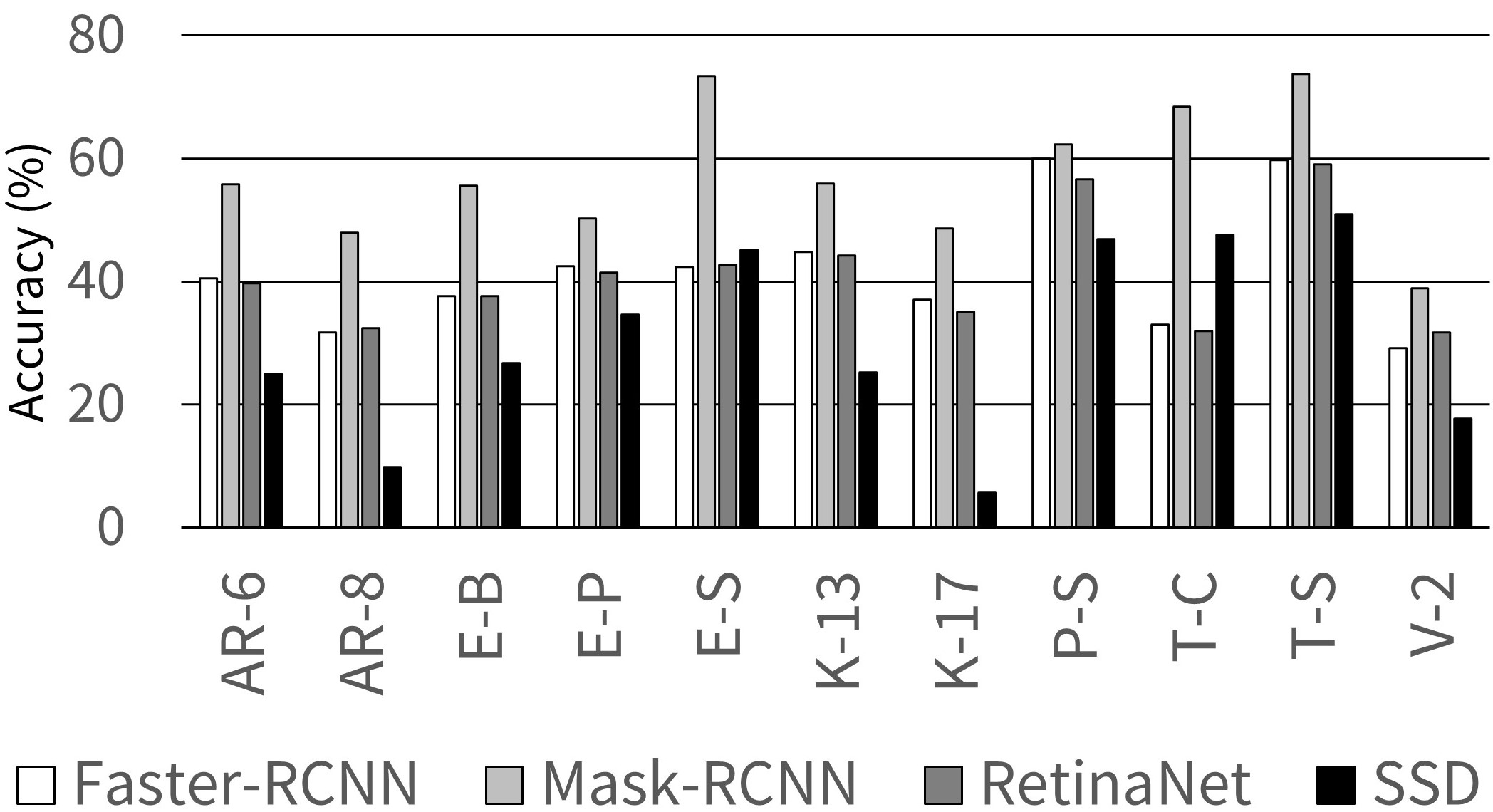}
    }
    \caption{
    \small Object detector consistency (our method, (a)) and accuracy (mAP, (b)) measured on the different videos in the MOT Challenge. Videos are:
     AR-6: ADL-Rundle-6,
    AR-8: ADL-Rundle-8, 
    E-B: ETH-Bahnhof, 
    E-P: ETH-Pedcross2, 
    E-S: ETH-Sunnyday, 
    K-13: KITTI-13, 
    K-17: KITTI-17, 
    P-S: PETS09-S2L1, 
    T-C: TUD-Campus, 
    T-S: TUD-Stadtmitte, 
    V-2: Venice-2. 
 State-of-the-art object detectors exhibit inconsistent behavior, ranging from 83.2\% to 97.1\% consistency. The two-stage models are both more consistent (a) and more accurate (b) than their single-shot counterparts. \normalsize}
    \label{fig:normal-consistency-results}
\end{figure}


As shown in Fig. \ref{fig:normal-consistency-results}, all object detectors exhibit some inconsistent behavior, ranging from 83.2\% to 97.1\% consistency ($C_V$ as calculated in Eqn. \ref{eqn:overall-consistency}). We also see that the two-stage models are more accurate and more consistent.

\section{Toward consistent object detection}
Object detector inconsistency is caused by the detector missing an object. As described in Related Work, missed detections can be caused by adversarial attacks and image distortions. Because the MOT Challenge is not an adversarial dataset, we expect the inconsistencies to be caused by image distortions naturally present in the dataset. Therefore, we apply different image distortion corrections from literature to compare their efficacies at improving consistency.

\begin{figure}
    \centering
    \begin{tcolorbox}[width=\linewidth]
        \section{Improving consistency via training}
        Common training techniques such as \textit{data augmentation} and \textit{dropout} are known to improve a model's robustness to image transformations and distortions. Despite these techniques being employed to produce our pretrained models, we find that inconsistencies still persist. To improve consistency, we use post-training image distortion corrections because of their accessibility. However, emerging training methods appear promising for improving consistency.  Zhang et al. propose weakly-supervised, context-based techniques~\cite{zhang_spftn_2020, zhang_leveraging_2019} that gather context for the scene (e.g., optical flow and prior knowledge) to provide additional information to an object detector - this information could stabilize the detector and improve consistency over time-series data from videos. Other techniques project neural network features into low-dimensional representations during training~\cite{dorsi_sparse_2020} - this could improve consistency on images of similar objects taken from different angles, lighting, etc.
    \end{tcolorbox}
\end{figure}


1) Gaussian Denoise (GD). Random sensor noise and shot noise can decrease detection performance. As demonstrated by Kang et al. \cite{kang_testing_2020}, we apply a normalized Gaussian filter to all images in the dataset in an attempt to reduce the noise.

2) Horizontal Flip (HF). Zhang, et al. \cite{zhang_making_2019} note that the slight translation of an object in an image could cause a previously misdetected object to become correctly detected. Further, Yin, et al. \cite{yin_defense_2020} observe that horizontally flipping an image can mitigate the detrimental effects of noise on a neural network. Thus, we horizontally flip all images.

3) WEBP Compression (WC). Compressing an image using the lossy JPEG format is already known to defeat adversarial attacks \cite{dodge_understanding_2016}. Yin, et al. \cite{yin_defense_2020} find that because WEBP compression introduced loop filtering, it is even better suited to breaking down the structures in an image that result from synthetic image distortions. Thus, we apply WEBP compression to the dataset images using a compression quality factor of 30 (Yin, et al. find that lower quality factors allow neural networks to perform better).

4) Unsharp Mask (UM). Tung, et al. \cite{tung_large-scale_2019} explain that if an object is moving, the motion blur can make it harder for a network to extract features along the object’s edges. The Unsharp Mask is a linear image processing technique commonly used to remove blur. The technique first identifies a set of blurry details by subtracting a further-blurred image from the original. The details are then emphasized in the original image. We apply the Unsharp Mask to the dataset to reduce the blur on object edges.

5) Gamma Correction (GC). Yeu, et al. \cite{yeu_investigation_2019} show that artificially increasing an image’s contrast and perceived brightness can help object detectors like Faster-RCNN perform better, particularly when the detectors are trained on daytime images. Gamma Correction is a common brighten/contrast technique that is driven by the Power Law expression; we use it on the dataset as well.

\begin{table}[h]
\caption{Avg. Consistency Improvements}
\begin{tabular}{|l|l|l|l|l|}
\hline
              & \begin{tabular}[c]{@{}l@{}}Faster-\\ RCNN\end{tabular} & \begin{tabular}[c]{@{}l@{}}Mask-\\ RCNN\end{tabular} & RetinaNet      & SSD            \\ \hline
GD             & 0\%                                                    & -0.3\%                                               & 0\%            & -0.6\%         \\ \hline
HF             & -5.3\%                                                 & -5.4\%                                               & -7.3\%         & -10.1\%        \\ \hline
\textbf{WC}    & \textbf{0.6\%}                                         & \textbf{0.5\%}                                       & \textbf{0.7\%} & \textbf{0.4\%} \\ \hline
\textbf{UM}    & \textbf{3.6\%}                                         & \textbf{2.6\%}                                       & \textbf{3.0\%} & \textbf{1.1\%} \\ \hline
\textbf{WC+UM} & \textbf{5.1\%}                                         & \textbf{3.0\%}                                       & \textbf{3.2\%} & \textbf{1.3\%} \\ \hline
GC             & 0.1\%                                                  & 0.1\%                                                & 0.4\%          & 0.1\%          \\ \hline
\end{tabular}
\label{tab:consistency-improvements}
\end{table}

\begin{table}[h]
\caption{Avg. Accuracy Improvements}
\begin{tabular}{|l|l|l|l|l|}
\hline
              & \begin{tabular}[c]{@{}l@{}}Faster-\\ RCNN\end{tabular} & \begin{tabular}[c]{@{}l@{}}Mask-\\ RCNN\end{tabular} & RetinaNet      & SSD            \\ \hline
GD             & 2.1\%                                                  & 2.4\%                                                & -0.6\%         & -1.1\%         \\ \hline
HF             & -19.3\%                                                & -19.4\%                                              & -25.5\%        & -28.4\%        \\ \hline
\textbf{WC}    & \textbf{1.5\%}                                         & \textbf{1.8\%}                                       & \textbf{0.5\%} & \textbf{0.5\%} \\ \hline
\textbf{UM}    & \textbf{2.0\%}                                         & \textbf{3.2\%}                                       & \textbf{8.3\%} & \textbf{3.6\%} \\ \hline
\textbf{WC+UM} & \textbf{3.2\%}                                         & \textbf{4.1\%}                                       & \textbf{8.6\%} & \textbf{3.9\%} \\ \hline
GC             & 0.1\%                                                  & -0.5\%                                               & -0.7\%         & -0.1\%         \\ \hline
\end{tabular}
\label{tab:accuracy-improvements}
\end{table}

Tab. \ref{tab:consistency-improvements} shows the average improvement across the MOT Challenge in terms of consistency percentage points (i.e. a table entry of Y\% means that it raises consistency from X\% to X+Y\%),  when the different distortion corrections are applied. Similarly, Tab. \ref{tab:accuracy-improvements} shows the accuracy improvement.

We see that both WEBP Compression (WC) and Unsharp Mask (UM) improve both consistency and accuracy for all object detectors. Applying both effects at the same time (WC+UM) gives a further overall improvement. In fact, the example inconsistencies in Fig. \ref{fig:maskrcnn-inconsistent} and \ref{fig:object-detectors-inconsistent} are resolved using WC+UM. Gaussian Denoise (GD) and Horizontal Flip (HF) both degrade consistency and accuracy (likely because applying these effects on relatively un-noisy images degrades the feature structure of the images \cite{zhang_fsim_2011}).

Finally, we note that improvements in consistency do not always equate to improvements in accuracy. Gamma Correction (GC) improves consistency, but degrades accuracy: in other words, the detector is consistently worse on the GC data, as described earlier in \autoref{fig:consistency-works-with-accuracy}.
\section{Conclusion and Future Work}
Object detectors are vital to many modern computer vision applications.
However, even state-of-the-art object detectors exhibit inconsistent behavior when the input undergoes small changes.
This inconsistent behavior is not fully captured by existing measurement tools; accuracy metrics and popular image datasets cannot measure whether the same objects are detected consistently.
We devise a consistency measurement method that uses images from videos and object ID labels.
Our method compliments accuracy measurement.
Using this method, we show that object detectors have consistency ranging from 83.2\% up to 97.1\%, depending on the input data.
Additionally, applying image distortion corrections like WEBP Compression and Unsharp Masking can improve consistency by as much as 5.1\%.
There is still room for improvement by the community. We only explore post-training methods to raise consistency. Future exploration should explore training-aware consistency improvements and further investigate the relationship between accuracy and consistency.

\section{Acknowledgments}
This project is supported in part by the National Science Foundation
OAC-2104709.
Any opinions, findings, and conclusions or recommendations expressed in this material are those of the author(s) and do not necessarily reflect the views of the National Science Foundation.

\section{Authors}
\textbf{Caleb Tung} is a doctoral student in the Elmore Family School of Electrical and Computer Engineering at Purdue University. Caleb’s research is on designing energy-efficient neural network operations for computer vision on embedded devices.
Outside the lab, Caleb can be found at the piano bench or somewhere in a good book.

\textbf{Abhinav Goel} is a doctoral candidate in the Elmore Family School of Electrical and Computer Engineering at Purdue University. Abhinav’s research interests are in the area of energy-efficient artificial intelligence systems. 

\textbf{Fischer Bordwell} is a Master’s student in the Elmore Family School of Electrical and Computer Engineering at Purdue University. Fischer’s research is on utilizing computer vision to reduce misinformation in the news. 

\textbf{Nick Eliopoulos} is currently is a doctoral student the Elmore Family School of Electrical and Computer Engineering at Purdue University. 
Outside of academics, Nick enjoys drawing and cooking.

\textbf{Xiao Hu} is an M.S. student in the Elmore Family School of Electrical and Computer Engineering at Purdue University. 
His research broadly connects to the fields of human-computer interaction, low-power computer vision, fairness in artificial intelligence, machine learning, and full-stack web development.

\textbf{George K. Thiruvathukal} is a professor and chairperson in the Department of Computer Science of Loyola University Chicago and visiting computer scientist at Argonne National Laboratory in the Leadership Computing Facility. His research interests include parallel computing, software engineering, and computer vision.

\textbf{Yung-Hsiang Lu} is a professor in the Elmore Family School of Electrical and Computer Engineering at Purdue University. He is also the director of Purdue's John Martinson Engineering Entrepreneurial Center. His research topics include computer systems, computer vision, and embedded systems. He is a Fellow of the IEEE and Distinguished Scientist of the ACM.
\bibliographystyle{ieeetr}
\bibliography{Multimedia2021}

\end{document}